\documentclass[journal]{IEEEtran}

\usepackage{amsmath,amssymb}
\usepackage{graphicx}
\usepackage{booktabs}
\usepackage{array}
\newcolumntype{L}[1]{>{\raggedright\arraybackslash}p{#1}}
\usepackage[hidelinks]{hyperref}
\usepackage{cite}
\usepackage{algorithm}
\usepackage{algpseudocode}

\newtheorem{remark}{Remark}

\newenvironment{standfirst}
  {\par\noindent\bfseries\ignorespaces}
  {\par\vspace{0.6\baselineskip}}

\newcommand{\R}{\mathbb{R}}
\newcommand{\norm}[1]{\lVert #1 \rVert}
\newcommand{\vs}{\mathbf{s}} \newcommand{\ve}{\mathbf{e}}
\newcommand{\vx}{\mathbf{x}} \newcommand{\vy}{\mathbf{y}}
\newcommand{\vJ}{\mathbf{J}} \newcommand{\vJh}{\hat{\mathbf{J}}}
\newcommand{\vq}{\mathbf{q}} \newcommand{\vr}{\mathbf{r}}
\newcommand{\vu}{\mathbf{u}}
\newcommand{\dq}{\Delta\vq}
\newcommand{\dQ}{\Delta\mathbf{Q}}
\newcommand{\dS}{\Delta\mathbf{S}}
\newcommand{\ds}{\Delta\vs}

\usepackage[most]{tcolorbox}

\newtcolorbox{runningexample}[1]{
    enhanced,
    breakable,
    colback=white,
    colframe=DiagBlue,
    boxrule=0pt,
    leftrule=1.4pt,
    arc=0pt,
    outer arc=0pt,
    left=6pt,
    right=2pt,
    top=4pt,
    bottom=4pt,
    before skip=6pt,
    after skip=6pt,
    before upper={
        \textcolor{DiagBlue}{\bfseries Running example: #1.}\quad
    },
}

\newtcolorbox{intuitions}[1]{
    enhanced,
    colback=white,
    colframe=IntuitionOrange,
    boxrule=0pt,
    leftrule=1.4pt,
    arc=0pt,
    outer arc=0pt,
    left=6pt,
    right=2pt,
    top=4pt,
    bottom=4pt,
    before skip=6pt,
    after skip=6pt,
    before upper={
        \textcolor{IntuitionOrange}{\bfseries Practical Interpretation: #1.}\quad
    },
}

\definecolor{DiagBlue}{RGB}{42,120,214}
\definecolor{IntuitionOrange}{RGB}{173,92,25}

\graphicspath{{figs/}}

\def\CodeURL{https://github.com/nachocz/tutorial_on_shape_servoing_of_deformable_objects}
\def\DemosURL{https://nachocz.github.io/tutorial_on_shape_servoing_of_deformable_objects/}
\newcommand{\CodeLink}{\expandafter\url\expandafter{\CodeURL}}
\newcommand{\DemosLink}{\expandafter\url\expandafter{\DemosURL}}
\newcommand{\linksragged}{\rightskip=0pt plus 3em\relax}
\usepackage[protrusion=false,expansion=true]{microtype}
\AtBeginDocument{\microtypesetup{expansion=false}}
\newcommand{\AEIcode}{\mbox{MICIU/}\allowbreak\mbox{AEI/}\allowbreak
  \mbox{10.13039/}\allowbreak\mbox{501100011033}}
\newcommand{\FundingAcknowledgment}{\section*{Acknowledgment}
{\microtypesetup{expansion=true}%
This work was supported by the REMAIN project S1/1.1/E0111 (Interreg Sudoe
Programme, ERDF); by project \mbox{PID2024-159279OB-I00}, funded by \AEIcode{}
and by ERDF/EU; by the Gobierno de Arag\'on, group T77\_26R; by grant
\mbox{RYC2024-051408-I}, funded by \AEIcode{} and by ESF+; by the Horizon
Europe project IntelliMan under grant number 101070136; by the SiMOD
project---Multi-robot Mobile Systems for Deformable Objects Manipulation (PR
FESR 21-27 Action 1.1.2, CUP J47G22000830003); and by the French National
Research Agency (ANR) under the France 2030 program with the reference
\mbox{ANR-25-PERO-0003} (PEPR Robotique -- PC DRMI). A.~Caporali is funded by
FSE+ (CUP J19J23000730002).\par}}

\begin{document}

\title{A Tutorial on Shape Servoing of Deformable Objects---Part~I:
Fundamentals}

\author{Ignacio~Cuiral-Zueco, Alessio~Caporali, Miguel~Aranda,\\
Gianluca~Palli, Gonzalo~L\'opez-Nicol\'as, and Youcef~Mezouar%
\thanks{This work has been submitted to the IEEE for possible publication.
Copyright may be transferred without notice, after which this version may no
longer be accessible.}%
\thanks{I.~Cuiral-Zueco, M.~Aranda, and G.~L\'opez-Nicol\'as are with the
Instituto de Investigaci\'on en Ingenier\'ia de Arag\'on (I3A), University of
Zaragoza, Spain. A.~Caporali and G.~Palli are with DEI, University of Bologna,
Italy. Y.~Mezouar is with Institut Pascal, Clermont Auvergne INP, France.
Corresponding author: I.~Cuiral-Zueco (e-mail: ignaciocuiral@unizar.es).}}
\markboth{Preprint version}%
{A Tutorial on Shape Servoing of Deformable Objects---Part~I}

\maketitle

\begin{standfirst}
In robotics, \textit{shape control}, or \textit{shape servoing}, refers to the
automatic manipulation of a deformable object toward a desired shape. It
involves a feedback loop in which the deformation Jacobian provides a local
linear approximation of how small commanded robot motions change the
measured shape of the object. Because the Jacobian depends on the object's
mechanical properties, geometry, and boundary conditions, deriving it accurately
from a mechanical model is often difficult. It can instead be estimated locally
from measurements collected during manipulation. This article is the first of a
two-part tutorial on this data-driven, Jacobian-based approach to shape control.
It introduces the components of the shape-servoing loop (shape features and
shape error, the deformation Jacobian, its offline estimation and online update,
and a standard control law), combines them into a complete controller, derives
a local stability condition, and explains why some target shapes cannot be reached.
Part~II covers more advanced topics. Throughout both parts, a simulated planar
example in which two robot arms manipulate an elastic rod observed by a camera
illustrates these concepts.
\end{standfirst}


\suppressfloats[t]
\begin{figure}[t]
    \centering
    \includegraphics[width=\linewidth]{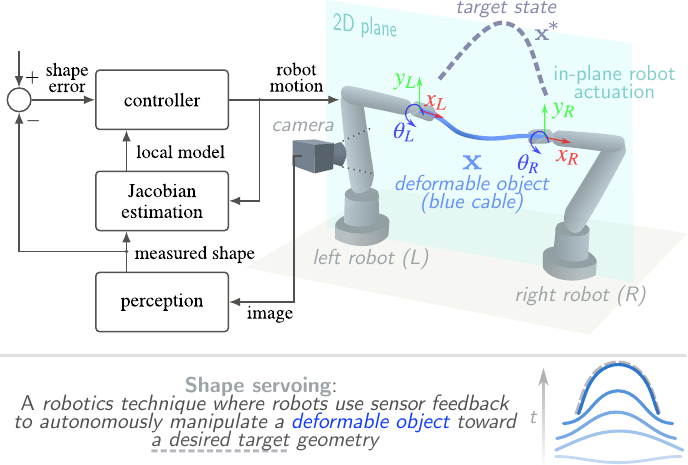}
    \caption{Illustration of a shape-servoing loop and of the setup (plant) used
    as the running example in this tutorial. Left: shape-servoing loop, with the
    measured shape obtained from the camera image and the shape error converted
    into robot motion using a local deformation model represented by the
    deformation Jacobian. The Jacobian is estimated from the applied robot
    motions and the corresponding measured shape changes. Right: running-example
    setup, in which two robot arms deform an elastic rod $\vx$ (blue), observed
    by a stationary camera, toward a target shape $\vx^{*}$ (dashed). For simplicity, in the running example, the grippers and the
    rod move in a plane parallel to the image plane of the camera, although this tutorial's formulation generalizes to three-dimensional
    actuation. The arrows at the grippers indicate the six in-plane actuation
    degrees of freedom $[{\color{red}x_L},{\color{green}y_L},{\color{blue}\theta_L},
 {\color{red}x_R},{\color{green}y_R},{\color{blue}\theta_R}]^{\top}$. Bottom: schematic
    definition of shape servoing, with an illustration of the object shape
    approaching the target over time $t$.}
    \label{fig:setup}
\end{figure}

\section{Introduction}
\label{sec:intro}

\subsection{Motivation and Purpose of This Tutorial}

\IEEEPARstart{D}{eformable} object manipulation is a growing area of research in
robotics, with applications in industrial automation, medical robotics, and
service robotics (e.g., cable routing, tissue manipulation, or cloth folding).
Many of these applications require regulating the shape of the object during
manipulation, for instance to achieve a desired configuration, avoid collisions,
or limit excessive stress. This task is known as \emph{shape control} or
\emph{shape servoing} (Fig.~\ref{fig:setup}). Although early works on shape
control emerged in the 2010s, the field has only recently matured to the point
where recurring methodological choices can be identified across a broad range of
applications. Building on these recurring choices, and as a natural extension of
the classical visual servoing tutorials
\cite{hutchinson1996,chaumette2006,chaumette2007}, this tutorial provides an
entry-level introduction to shape control. It focuses on \emph{data-driven
Jacobian-based shape control}, an approach that has become standard in the
literature.

Existing surveys and taxonomies give a broad overview of deformable-object
modeling, perception, learning, and manipulation
\cite{arriola2020,yin2021,cuiral2024,caporali2026robotic}, but they do not
provide a unified, control-oriented introduction to shape servoing. In
particular, the recurring Jacobian-based formulations are dispersed across
application-specific works and are often presented with different notation,
assumptions, and estimation strategies. This tutorial addresses that gap by
organizing these methods in a common framework and presenting the concepts
needed to understand, implement, and analyze data-driven Jacobian-based shape
control.

Table~\ref{tab:roadmap} summarizes the main concepts developed in Part~I. The
\textbf{accompanying supplementary material} provides \textbf{public code} and
\textbf{interactive demos},\footnote{\linksragged Code: \CodeLink. Interactive
demos: \DemosLink.} which implement the running example (Fig.~\ref{fig:setup})
and reproduce the numerical results shown in this paper.

\begin{table}[!t]
\caption{Contents of Part~I}
\label{tab:roadmap}
\centering
\renewcommand{\arraystretch}{1.2}
\footnotesize
\setlength{\tabcolsep}{4pt}
\begin{tabular}{@{}L{1.55cm}L{6.15cm}@{}}
\toprule
\textbf{Section} & \textbf{Main content} \\
\midrule
Section~\ref{sec:components} & Shape representation and shape error with respect to
the target, the
deformation Jacobian relating robot motion to shape change, and the control law
mapping shape error to robot motion. \\
Section~\ref{sec:estimation} & Estimation of the initial Jacobian from
exploratory motions, online adaptation from closed-loop data, and model-based
alternatives. \\
Section~\ref{sec:loop} & Integration of the components into a complete feedback
loop, characterization of reachable target shapes, and conditions for local
decrease of the shape error. \\
\bottomrule
\end{tabular}
\end{table}

\subsection{Why Data-Driven Jacobian-Based Shape Control?}

Shape control requires a local description of how robot motions affect the
measured shape of a deformable object \cite{cuiral2024}. This input--output
relation is captured by the \emph{interaction Jacobian}, referred to here as the
\emph{deformation Jacobian}. The Jacobian generally depends on the object's
mechanical properties (e.g., stiffness, density), geometry (e.g., surface
curvature, at-rest shape), and boundary conditions (e.g., grasping points,
gravity).

One option is to derive the deformation Jacobian from a mechanical or geometric
model. Such models can provide a physically meaningful description of the
deformation response, but may be computationally expensive and require object
properties, boundary conditions, or model parameters that are difficult to
measure or identify \cite{arriola2020,shetab2022,mcconachie2018}.

Alternatively, the Jacobian can be estimated directly from measurements
collected during manipulation and updated online as the object deforms. This
avoids explicitly modeling the object's mechanics while providing the local
input--output relation required for feedback control. In this tutorial,
\emph{model-free} refers specifically to this setting: no explicit mechanical or
geometric model is used, although the estimated Jacobian itself constitutes a
local input--output model \cite{cuiral2024}. This formulation is closely related
to uncalibrated visual servoing \cite{hosoda1994} and has become a standard
approach in shape control, with applications to cables
\cite{navarro2013,lagneau2020}, contours and surfaces
\cite{berenson2013manipulation,navarro2018}, and plastically
deformable materials \cite{cherubini2020}.

We adopt this data-driven formulation because it is well established in the
literature, compact enough to present a complete shape-servoing methodology
within a tutorial, and sufficiently modular to provide a common framework that
readers can first learn as a whole and then extend by studying or replacing
individual components. For example, as discussed above, the deformation Jacobian
may instead be obtained from a mechanical or geometric model \cite{arriola2020},
while alternative shape features or feedback laws can be used without changing
the overall control structure presented here.

\subsection{Application Scope of This Tutorial}
\label{sec:scope}

The methods in this tutorial apply to manipulation tasks that satisfy three
assumptions:

\begin{itemize}
    \item \emph{Quasistatic, velocity-level manipulation.} The object deforms
    slowly and \textit{only} in response to robot motion, with negligible
    inertial effects. Its deformation can therefore be treated as a sequence of
    quasistatic mechanical equilibrium configurations driven by robot velocity
    commands.

    \item \emph{Measurable shape feedback.} The deformation relevant to the task
    can be measured during manipulation, e.g., from vision or other sensing
    modalities. This may correspond to the full object shape or to a set of
    shape features. No particular perception pipeline or representation is
    required, provided that the selected features capture the deformation
    relevant to the control objective.

    \item \emph{Continuous manipulation regime.} The relation between robot
    motion and object deformation changes smoothly during the task. Abrupt
    changes in contact conditions (e.g., regrasping or collisions), changes in
    topology (e.g., breaking the object or attaching parts of it to form loops),
    or transitions between markedly different deformation regimes (e.g., in
    highly elastoplastic objects) may invalidate the current deformation
    Jacobian and require re-estimation or additional task-level logic.
\end{itemize}

Beyond these assumptions, the general framework presented here
(Fig.~\ref{fig:setup}) applies across different types of deformable objects,
sensing modalities, and methods for obtaining the deformation Jacobian.

\begin{figure*}[t]
\centering
\includegraphics[width=\textwidth]{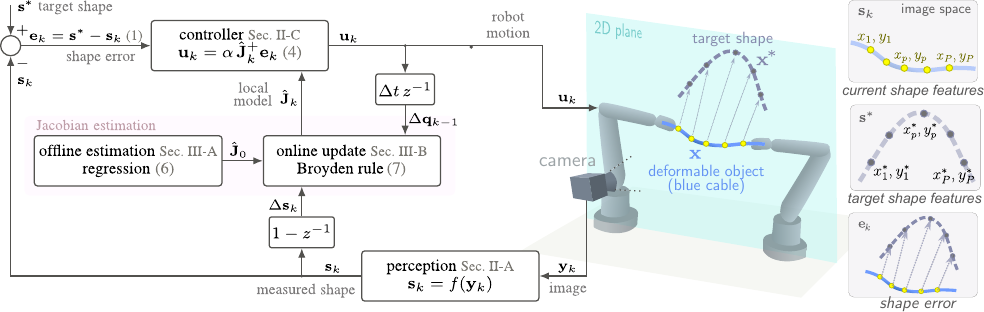}
\caption{Discrete-time shape-servoing loop of the running example, with the
sections and equations defining each block. Perception extracts the image
coordinates $\vs_k$ of the tracked points from the camera measurement $\vy_k$,
and the controller computes the command $\vu_k$ from the shape error $\ve_k$.
The operators $\Delta t\,z^{-1}$ and $1-z^{-1}$, where $z^{-1}$ denotes a
one-iteration delay, compute the latest robot-state increment $\dq_{k-1}$ and
feature change $\ds_k$. These are used by the online update \eqref{eq:broyden} to
correct the Jacobian estimate, initialized from the offline estimate \eqref{eq:regression} $\vJh_0$
(Fig.~\ref{fig:offline}). Right: current features $\vs_k$ (yellow), target
features $\vs^{*}$, and shape error $\ve_k$ in the image; the error arrows are
also shown on the rod in the scene.}
\label{fig:loop}
\end{figure*}
\section{Components of a Shape-Servoing Loop}
\label{sec:components}

A standard shape-servoing loop contains three main blocks (Fig.~\ref{fig:setup},
left): \textbf{perception}, which extracts shape features from sensor
measurements; \textbf{local model estimation}, which relates robot motion to
changes in those shape features; and the \textbf{feedback control law}, which
uses the measured shape features and estimated local model to generate motions
that reduce the shape error. In the detailed loop of Fig.~\ref{fig:loop}, these
correspond to the perception block, the offline estimation and online update
blocks, and the controller block, respectively. This section introduces the
quantities and operations associated with these blocks and illustrates them
using the planar rod running example.

\subsection{Shape Features and Shape Error}
\label{sec:errorspace}

Let $\vx$ denote the full configuration of the deformable object, e.g., a curve
for a cable (Fig.~\ref{fig:setup}), a surface for a cloth, or a closed surface
for a balloon. Sensors provide measurements $\vy(t)$ of this configuration, such
as tracked image points, a segmented contour, or a point cloud. A feature map
$\vs = f(\vy) \in \R^{n}$ extracts the finite-dimensional shape representation
used for control, such as the coordinates of sampled points, curvature values
along a contour, or modal coefficients (e.g., Fourier coefficients) that describe
the overall shape. The controller uses $\vs$, i.e., the reduced description of
$\vx$, as its feedback signal.

Given a target feature vector $\vs^{*}$, the shape error is
\begin{equation}
    \ve(t) = \vs^{*} - \vs\big(\vy(t)\big).
    \label{eq:error}
\end{equation}
For example, if $\vs$ stacks the coordinates of sampled points along a cable,
$\ve$ measures how far each sampled point is from its target location. This
requires each element of $\vs$ to retain a consistent physical meaning: the
$i$th element must track the same material point as the cable deforms, and the
$i$th element of $\vs^{*}$ must represent the target location of that same
material point \cite{cuiral2025time}. Other representations, such as modal
descriptors \cite{yang2023,navarro2018}, do not associate each feature with a
specific material point, but still require a consistent parameterization of the
curve or domain between the current and target shapes
\cite{cuiral2023shapecompliant}.

\begin{runningexample}{Shape Features}
The insets on the right of Fig.~\ref{fig:loop} illustrate this representation in
image space. Perception tracks $P=5$ material points along the rod (yellow dots,
top inset) and stacks their image coordinates as
$\vs=[x_1,y_1,\dots,x_P,y_P]^{\top}$, giving $n=10$. The target
$\vs^{*}\in\R^{10}$ contains the coordinates of the same material points on the
target shape (middle inset). Each arrow in the bottom inset, also shown on the
rod in the scene, represents the error of one point, corresponding to two
entries (2D) of $\ve_k$. Note that, in our example, camera calibration is not
required because the mapping from robot motion to changes in the shape features
is estimated directly from image measurements (Section~\ref{sec:estimation}).
\end{runningexample}

\begin{figure*}[t]
\centering
\includegraphics[width=\textwidth]{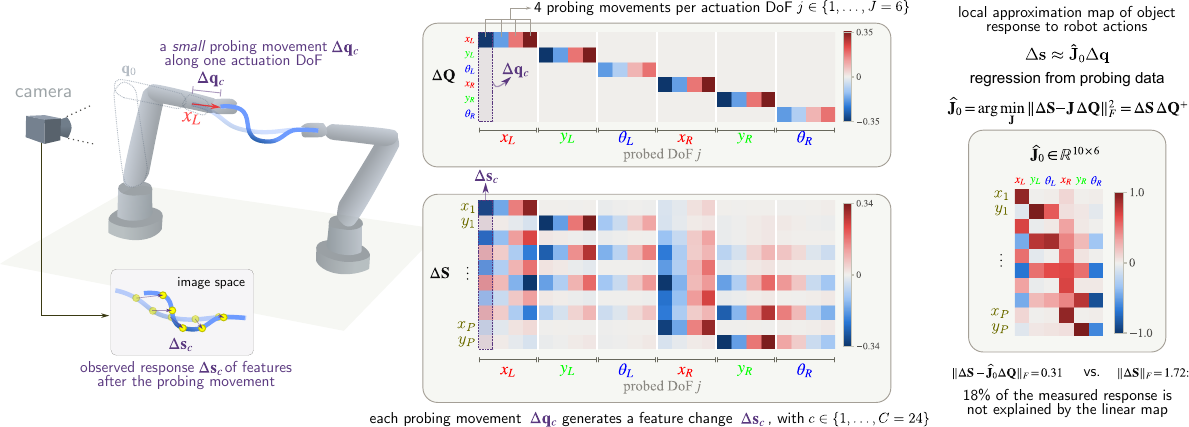}
\caption{Offline Jacobian estimation in the running example. Left: a small robot
perturbation $\dq_c$ produces a measured feature change $\ds_c$. Center: the
$C=24$ action--response pairs are stacked as the columns of $\dQ$ and $\dS$.
Right: least-squares regression gives the initial Jacobian estimate $\vJh_0$.}
\label{fig:offline}
\end{figure*}
\subsection{Quasistatic Actuation and the Deformation Jacobian}
\label{sec:plant}

Following the first assumption of Section~\ref{sec:scope}, we treat manipulation
as quasistatic: after a small robot motion, the object settles to a nearby
equilibrium before the next control action, so inertial effects are neglected.
This assumption excludes regimes dominated by fast dynamics\footnote{Slow and
quasistatic dynamics, like gravity effects in equilibrium configurations, can
still be addressed with the proposed data-driven approach.}, like abrupt
mechanical transitions, such as hysteresis or bifurcation events (e.g.,
buckling). To describe how the object's equilibrium shape depends on the robot
configuration, let $\vq\in\R^{m}$ collect the actuated robot states, such as
joint state, or end effector pose. For non-varying interaction conditions (e.g.,
same grasping points throughout the process), each robot configuration near the
current one determines a nearby equilibrium shape, so the features can be
written as $\vs = \vs(\vq)$.

The shape controller outputs a velocity command $\vu\in\R^{m}$ for the actuated
robot coordinates $\vq$. Assuming the low-level robot controller tracks this
command on a faster time scale than the shape-control loop, the robot actuation
can be approximated by the pure integrator $\dot{\vq} = \vu\in\R^{m}$.

\begin{runningexample}{Actuation and Simulation}
Both grippers move in the plane shown in Fig.~\ref{fig:setup} (right), each with
three degrees of freedom (DoF): two planar translations (${\color{red}x}$ and
${\color{green}y}$) and one rotation (${\color{blue}\theta}$), as indicated by
the arrows. Hence,
\[
\vq=
[{\color{red}x_L},{\color{green}y_L},{\color{blue}\theta_L},
 {\color{red}x_R},{\color{green}y_R},{\color{blue}\theta_R}]^{\top},
\qquad m=6.
\]
The rod is modeled as a Kirchhoff rod with stretching and bending stiffness,
discretized into $25$ nodes and subject to gravity. For each gripper
configuration, its quasistatic equilibrium is computed by minimizing the rod's
total potential energy.
\end{runningexample}

Because the shape features depend on the object configuration, which in turn
depends on the robot coordinates, the chain rule gives
$\dot{\vs}=(\partial\vs/\partial\vx)(\partial\vx/\partial\vq)\,\dot{\vq}$. The
factor $\partial\vx/\partial\vq$ describes how robot motion changes the object
configuration, and $\partial\vs/\partial\vx$ describes how those changes are
reflected in the shape features. Their product is the deformation Jacobian
\cite{navarro2013,lagneau2020},
\begin{equation}
    \vJ(\vq)
    =
    \frac{\partial \vs}{\partial \vx}
    \frac{\partial \vx}{\partial \vq}
    =
    \frac{\partial \vs}{\partial \vq}
    \in \R^{n\times m}.
    \label{eq:chain}
\end{equation}
With $\dot{\vq}=\vu$, the local input--output relation becomes
\begin{equation}
    \dot{\vs}
    =
    \vJ(\vq)\dot{\vq}
    =
    \vJ(\vq)\vu.
    \label{eq:fwd}
\end{equation}
The $j$th column of $\vJ$ is the feature velocity produced by a unit velocity of
robot coordinate $q_j$. As in visual servoing, $\vJ$ maps robot motion to
feature motion \cite{chaumette2006}. Unlike in visual servoing, however, this
mapping is not determined solely by the camera projection model and feature
geometry, but also by the object's configuration-dependent mechanical response.
The same gripper motion can therefore produce very different shape changes, for
example when a rod is slack versus taut (Part~II).

\begin{intuitions}{Deformation Jacobian}
The deformation Jacobian is a matrix that describes how small robot motions
change the object's shape. It can be estimated directly from real data, using
measured robot motions and the shape changes they produce. As the object
deforms, the same robot motions can produce different shape changes, so the
Jacobian must be updated during manipulation.
\end{intuitions}

\begin{runningexample}{Deformation Jacobian}
The Jacobian $\vJ\in\R^{10\times6}$ maps the $m=6$ velocities of grippers
(Fig.~\ref{fig:setup}) to the velocities of $n=10$ image coordinates of the
tracked points in Fig.~\ref{fig:loop}. The right panel of Fig.~\ref{fig:offline}
displays its estimate as a $10\times6$ color map, with one column per gripper
coordinate and one row per image coordinate.
\end{runningexample}

\subsection{Control Law}
\label{sec:control}

The true Jacobian $\vJ$ is generally unknown, so the controller uses a
\textit{sufficiently accurate} estimate $\vJh \approx \vJ$, obtained from data or from a model of the object
(Section~\ref{sec:estimation}). Given $\vJh$, the controller block in
Fig.~\ref{fig:loop} maps the shape error \eqref{eq:error} to a robot velocity
using the Moore--Penrose pseudoinverse\footnote{Near rank loss, the
pseudoinverse becomes ill-conditioned. Part~II introduces the damped inverse used in that case.}, denoted by $(\cdot)^{+}$:
\begin{equation}
    \vu = \alpha\,\vJh^{+}\ve,
    \label{eq:ctrl}
\end{equation}
where $\alpha>0$ is the control gain. Ideally, the features would then move with
the desired velocity $\alpha\ve$, and the error would decay exponentially. With
$\vJh$ in place of $\vJ$ in \eqref{eq:fwd}, this command yields the predicted feature velocity
$\vJh\vu=\alpha\,\vJh\vJh^{+}\ve$. This is the orthogonal projection of
$\alpha\ve$ onto $\operatorname{Range}(\vJh)$: in other words, the closest
feature velocity to $\alpha\ve$ that the estimated model can produce. Here,
$\operatorname{Range}(\vJh)$ denotes the set of all feature velocities
obtainable from some robot velocity. The control law therefore acts only on the
part of the shape error that can be produced through robot motion under the
estimated model.

When $n>m$ and $\vJh$ has full column rank, meaning that its columns are linearly
independent, as in the running example,
$\vJh^{+}=(\vJh^{\top}\vJh)^{-1}\vJh^{\top}$, and \eqref{eq:ctrl} yields the
robot velocity that minimizes the least-squares error in feature space, i.e.,
makes the predicted feature velocity \textit{as close as possible} to the
desired one. When $n<m$ and $\vJh$ has full row rank, meaning that its rows are linearly
independent (e.g., for the curvature representation of Part~II of this
tutorial), $\vJh^{+}=\vJh^{\top}(\vJh\vJh^{\top})^{-1}$, and \eqref{eq:ctrl}
yields the minimum-norm robot velocity: among all robot velocities that realize
the desired feature velocity, it \textit{selects} the one with the smallest
magnitude. In the latter case, the null space $\operatorname{Null}(\vJh)$ is
nontrivial and contains robot velocities that $\vJh$ maps to zero. Intuitively,
these ``extra'' robot motions that do not change the selected shape features can
be used for other tasks or control objectives without altering the predicted
shape motion.

\subsection{Discrete-Time Implementation}
\label{sec:discrete}

The preceding formulation is given in continuous time, whereas the control loop
typically operates at discrete iterations $k$ with period $\Delta t$
(Fig.~\ref{fig:loop}). At iteration $k$, perception provides features
$\vs_k=f(\vy_k)$, after which the controller evaluates \eqref{eq:error} and
\eqref{eq:ctrl} as $\ve_k=\vs^{*}-\vs_k$ and the control law \eqref{eq:ctrl}
becomes
\begin{equation}
    \vu_k=\alpha\,\vJh_k^{+}\ve_k.
    \label{eq:uklaw}
\end{equation}

The robot then applies $\vu_k$ over one control period, producing the coordinate
increment $\dq_k=\vu_k\Delta t$. Consistent with the quasistatic assumption, the
object mechanically settles within each period, such that each measurement
$\vs_k$ corresponds to an equilibrium shape.

The latest produced robot increment $\dq_{k-1}=\vu_{k-1}\Delta t$ and its
resulting feature change $\ds_k=\vs_k-\vs_{k-1}$ form an \emph{action--response
pair}. For small increments, \eqref{eq:fwd} gives $\ds_k\approx\vJ\dq_{k-1}$,
the relation from which Section~\ref{sec:estimation} estimates and updates the
Jacobian. In Fig.~\ref{fig:loop}, the operators $\Delta t\,z^{-1}$ and
$1-z^{-1}$, where $z^{-1}$ denotes a delay of one iteration, form this pair from
$\vu_k$ and $\vs_k$, and the online update block returns the estimate $\vJh_k$
to the controller.

\section{Estimating the Deformation Jacobian}
\label{sec:estimation}

The control law \eqref{eq:uklaw} requires an initial Jacobian estimate $\vJh_0$
(the subscript $0$ refers to $k=0$) and a way to keep the estimate
representative of the object's response as it deforms ($k>0$). In
Fig.~\ref{fig:loop}, the offline estimation block obtains $\vJh_0$ by regression
from small exploratory motions (Section~\ref{sec:offline}), while the online
update block adapts the estimate during control using newly recorded data
through a secant update (Section~\ref{sec:online}). Alternatively, a model of
the object can provide the Jacobian (Section~\ref{sec:modelbased}).

\subsection{Offline Estimation by Regression}
\label{sec:offline}

To initialize the Jacobian, the robots apply $C$ small exploratory motions
around their initial configuration $\vq_0$, and the resulting feature change
from each motion is measured (Fig.~\ref{fig:offline}). For exploratory motion
$c$, define the \emph{action--response pair} $(\dq_c,\ds_c)$ as the
corresponding robot-state and feature increments: $\dq_c=\vq_c-\vq_0$ and
$\ds_c=\vs_c-\vs_0$, respectively. These increments are stacked as columns of
$\dQ\in\R^{m\times C}$ and $\dS\in\R^{n\times C}$, respectively. For
sufficiently small motions, $\ds_c\approx\vJ\dq_c$ (Section~\ref{sec:discrete}),
so the initial Jacobian is estimated by a least-squares regression over all
measured pairs:
\begin{equation}
    \vJh_0
    =
    \arg\min_{\vJ}
    \big\lVert \dS-\vJ\dQ \big\rVert_F^2
    =
    \dS\dQ^{+},
    \label{eq:regression}
\end{equation}
where $\norm{\cdot}_F$ denotes the Frobenius norm. If the exploratory motions span the input space, meaning that together they
excite all independent directions in which the robots can move, $\dQ$ has full
row rank and the least-squares solution is unique. In this tutorial, we ensure
this by perturbing each actuation DoF independently, one at a time. More general
exploratory motions can also be used, provided that their action increments
collectively span the input space, i.e., $\operatorname{rank}(\dQ)=m$, which
requires at least $C\geq m$ motions.

The probing amplitude must be small enough for the first-order approximation
$\ds_c\approx\vJ\dq_c$ to remain valid, yet large enough for the resulting
feature changes to exceed measurement noise. The regression residual
$\norm{\dS-\vJh_0\dQ}_F$ therefore quantifies how well a single linear map
explains the collected responses. A small residual, however, only supports the
estimate locally around the probed configuration; as the object deforms, the
Jacobian estimate must be updated.

\begin{runningexample}{Offline Jacobian Estimation}
The two grippers have three DoFs each, for a total of six actuation DoFs. For
each DoF, we apply four exploratory motions: two in the positive direction and
two in the negative direction, chosen so that the robots return to their initial
state $\vq_0$ after a DoF is probed. This yields $C=24$ action--response pairs.
The left panel of Fig.~\ref{fig:offline} shows one such perturbation along the
${\color{red}x_L}$ direction and the resulting feature displacement $\ds_c$ in
image space. In the center panel, the perturbations are assembled in
$\dQ\in\R^{6\times24}$, with each column exciting only one DoF, while the
corresponding measured responses form the columns of $\dS\in\R^{10\times24}$.
The right panel shows the fitted Jacobian $\vJh_0$, whose residual
$\norm{\dS-\vJh_0\dQ}_F\approx0.31$ is approximately $18\%$ of
$\norm{\dS}_F\approx1.72$; that is, a single linear map leaves about $18\%$ of
the measured response unexplained.
\end{runningexample}

A singular value decomposition (SVD) of $\vJh_0$ characterizes how strongly
robot motion changes the shape along each direction. It pairs robot-motion
directions with the feature directions they produce, each with a gain given by a
singular value $\sigma_l$. Because the pseudoinverse in \eqref{eq:ctrl} divides
by these gains, small singular values can turn small feature errors into large
robot commands. The condition number $\sigma_{\max}/\sigma_{\min}$, the ratio of
the largest to the smallest singular value, summarizes this imbalance; in the
running example, $\vJh_0$ has full column rank and a condition number of about
$17$. Part~II of this tutorial develops the SVD analysis further and uses it
to explain and interpret how the shape servoing loop can fail.

\subsection{Online Update: Broyden's Rule}
\label{sec:online}

As the object deforms, the initial estimate $\vJh_0$ may no longer represent its
current local response. Re-estimating the Jacobian by repeating the offline
probing procedure of Section~\ref{sec:offline} would require interrupting the
control process. Instead, the estimate can be updated from data already
generated during closed-loop control: each control action $\dq_{k-1}$ from
\eqref{eq:uklaw} produces an observed feature change $\ds_k$. These
action--response pairs can therefore be used directly to update the Jacobian
online, without additional probing, as represented by the online update block in
Fig.~\ref{fig:loop}. For the latest step, the prediction error of the current
estimate is $\vr_k=\ds_k-\vJh_{k-1}\dq_{k-1}$, and Broyden's rule
\cite{broyden1965} corrects the estimate as
\begin{equation}
    \vJh_k
    =
    \vJh_{k-1}
    +
    \frac{\vr_k\,\dq_{k-1}^{\top}}
         {\dq_{k-1}^{\top}\dq_{k-1}}.
    \label{eq:broyden}
\end{equation}
The update makes the new estimate exactly reproduce the latest observed
action--response pair, $\vJh_k\dq_{k-1}=\ds_k$, while changing $\vJh_{k-1}$ as
little as possible in Frobenius norm.

\begin{intuitions}{Using Control Motions and Measurements to Update the Jacobian}
During the control process, each commanded robot motion provides new data about
how the object deforms. The current Jacobian estimate can be used to predict the
resulting change in shape features. This prediction is then compared with the
feature change measured by perception, yielding a prediction error. Broyden's
rule uses this error to update the Jacobian estimate as manipulation proceeds.
\end{intuitions}

The same idea appears in uncalibrated visual servoing \cite{hosoda1994} and in
model-free shape servoing papers such as \cite{navarro2013,lagneau2020}.

\begin{remark}[Gating small updates]
\label{rem:gate}
The magnitude of the correction in \eqref{eq:broyden}, measured in the Frobenius
norm, is $\norm{\vr_k}/\norm{\dq_{k-1}}$, where $\norm{\cdot}$ denotes the
Euclidean norm of a vector. Under smooth model variation, the prediction error
typically decreases with the commanded step. Measurement noise, however, does
not decrease with the step size, so dividing by a very small command can produce
an excessively large update. Practical implementations therefore skip updates
when $\norm{\dq_{k-1}}$ falls below a minimum threshold and reject corrections
that are implausibly large.
\end{remark}

\subsection{Model-Based and Hybrid Estimation}
\label{sec:modelbased}

The Jacobian estimate can also be obtained from a mechanical, geometric, or
learned model of the object and the robot, either at every iteration or as an
initial estimate refined online by the Broyden update. Depending on the model, the Jacobian
may be derived analytically or estimated in simulation by replicating the
offline Jacobian estimation procedure of Section~\ref{sec:offline}. These models can
also be refined online from real data collected during control, for example
through parameter estimation. Examples include as-rigid-as-possible models
\cite{shetab2022}, finite-element models, and mass--spring models.

The main drawback of such model-based approaches is that relating the model to
the physical setup may require camera calibration, the object's at-rest
geometry, sufficiently accurate initial mechanical parameters, grasp locations
and other constraints, and explicit knowledge of the available actuation DoFs.
Errors in any of these elements can also affect the Jacobian estimate. The
model-free approach from this tutorial avoids these requirements: it does not
need an explicit object model, physical parameters, camera--robot calibration,
or prior knowledge of the grasp configuration. In either case, closed-loop
behavior depends on how well $\vJh$ represents the current local response
(Section~\ref{sec:stability}).

\section{Closed-Loop Shape Servoing}
\label{sec:loop}
This section assembles the components of
Sections~\ref{sec:components}--\ref{sec:estimation} into the complete
shape-servoing loop of Fig.~\ref{fig:loop}. It then separates two questions that
are important in practice: whether a target can be realized and reached by the
manipulation system, and whether the local feedback law converges once the
system is near that target.

\begin{figure}[t]
    \centering
    \includegraphics[width=\columnwidth]{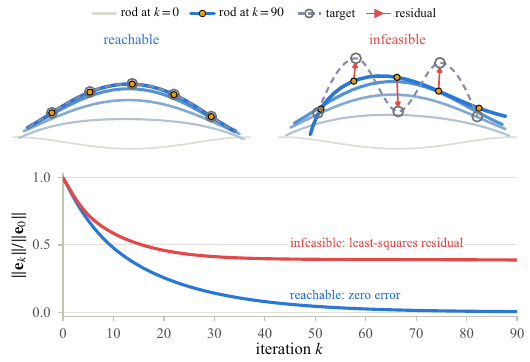}
    \caption{Targets with different closed-loop outcomes in the running example.
    Top: rod configurations from $k=0$ (light grey) to $k=90$ (blue, with yellow
    feature points) for two target shapes. Left: a target recorded at another
    gripper configuration, which is reached from the tested initial state.
    Right: a modified target obtained by displacing three feature points, for
    which the controller settles with a nonzero residual. Dashed curves with
    grey rings indicate the target shapes, and red arrows denote the residual
    feature error. Bottom: normalized error histories for the two cases.}
    \label{fig:reachability}
\end{figure}

\begin{figure*}[t]
    \centering
    \includegraphics[width=\textwidth]{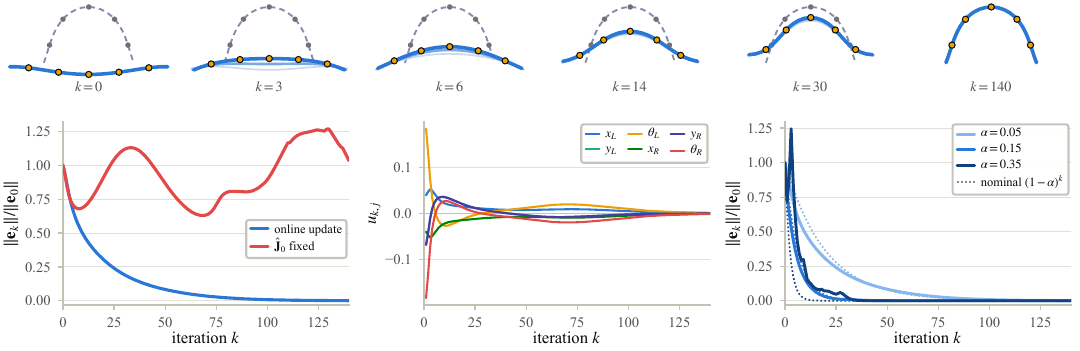}
    \caption{Closed-loop behavior for the arch target of Fig.~\ref{fig:setup}.
    Top: rod configurations (blue, with yellow feature points) and dashed target
    at the indicated iterations $k$, selected at approximately equal intervals
    of travel in feature space; lighter curves show two preceding iterations.
    Bottom, from left to right: normalized error with online Jacobian updates
    and with $\vJh_0$ held fixed ($\alpha=0.05$); commanded robot velocities
    ($\alpha=0.05$; $y_L$ and $y_R$ coincide by symmetry); and normalized error for three control gains (solid), each with its
    nominal decay $(1-\alpha)^k$ (dotted, same shade).}
    \label{fig:control}
\end{figure*}

\subsection{Complete Shape-Servoing Loop}
\label{sec:closedloop}

Algorithm~\ref{alg:servoing} illustrates the steps of the loop shown in
Fig.~\ref{fig:loop}. Starting from the first shape measurement and the initial
estimate $\vJh_0$, at each iteration the controller computes the shape error and
the control action \eqref{eq:uklaw}. The robot applies the command for one
control period, after which the resulting shape change is measured and used to
update the Jacobian estimate through \eqref{eq:broyden}. The loop runs until a
task-level supervisor stops it, for example when $\norm{\ve_k}$ falls below a
chosen tolerance.

\begin{algorithm}[t]
\caption{Online Jacobian-based shape servoing}
\label{alg:servoing}
\begin{algorithmic}[1]
\Require target $\vs^{*}$, initial estimate $\vJh_0$, gain $\alpha$, period
         $\Delta t$ \State measure $\vy_0$ and compute $\vs_0=f(\vy_0)$; \ $k
         \gets 0$ \While{true} \State $\ve_k \gets \vs^{*}-\vs_k$ \State $\vu_k
         \gets \alpha\,\vJh_k^{+}\ve_k$ \Comment{control law \eqref{eq:uklaw}}
         \State apply $\vu_k$ for $\Delta t$; \ $\dq_k \gets \vu_k\Delta t$
         \State measure $\vy_{k+1}$ and compute $\vs_{k+1}=f(\vy_{k+1})$ \State
         $\ds_{k+1} \gets \vs_{k+1}-\vs_k$ \State $\vJh_{k+1} \gets$ update of
         $\vJh_k$ from $(\dq_k,\ds_{k+1})$ \Comment{\eqref{eq:broyden}} \State
         $k \gets k+1$ \EndWhile
\end{algorithmic}
\end{algorithm}

Recall that $\ds_{k+1}$ is the measured response to the robot command $\dq_k$.
This action--response pair updates the Jacobian estimate, which is then used to
compute the next control command. In practice, line~8 of
Algorithm~\ref{alg:servoing} leaves $\vJh_{k+1}=\vJh_k$ unchanged when
$\norm{\dq_k}$ is very small (Remark~\ref{rem:gate}).

\subsection{Target Feasibility and Reachability}
\label{sec:targets}

Not every target shape can be obtained with a given manipulation setup,
regardless of which control law is being used. The robot motion and workspace,
the grasping configuration, the physical deformation limits, and equilibria that
the object can sustain all constrain the set of achievable shapes.

In control terms, a target is \emph{feasible} if it belongs to the set of
admissible states, and \emph{reachable} from a given initial state if an
admissible input trajectory can drive the system to such a target state. For the
shape-servoing problem considered here, let $\mathcal{S}\subseteq\R^n$ denote
the set of feature vectors realizable under a particular robot setup, deformable
object, and quasistatic assumptions introduced in Section~\ref{sec:scope}. A
target $\vs^*$ is feasible if $\vs^*\in\mathcal{S}$, and reachable from the
current configuration if an admissible manipulation path exists to a
configuration that realizes $\vs^*$.

Feasibility and reachability are global properties, whereas the deformation
Jacobian used by the controller is a local model. From \eqref{eq:fwd}, the
instantaneous feature velocities are restricted to $\operatorname{Range}(\vJ)$,
whose dimension is $\operatorname{rank}(\vJ)\leq\min(n,m)$. Hence, the Jacobian
characterizes which shape variations are locally reachable at the current
configuration, but it cannot by itself establish whether a target is globally
reachable from the current configuration through an admissible manipulation
path. This same locality also defines the scope of the convergence and stability
analysis of the closed loop, considered next.

\begin{runningexample}{Feasibility, Reachability and Residual Error}
The running example has $n=10$ feature dimensions and $m=6$ actuation DoFs, so
at most six independent feature-velocity directions are available locally.
Figure~\ref{fig:reachability} compares two targets from the same initial state.
The first target was recorded at another admissible gripper configuration and
thus is feasible by construction. Convergence to zero error during control also
demonstrates that this target is reachable from the given initial state. The
second target is obtained by modifying the first target, displacing three target
feature points beyond the object's admissible deformation limits. It is
therefore considered infeasible.
\end{runningexample}

\subsection{Local Stability and Convergence}
\label{sec:stability}

To analyze local convergence, we assume a fixed target shape, such that
$\dot{\vs}^{*}=\mathbf{0}$. We first consider the ideal case in which the
deformation Jacobian is known exactly, so that the controller uses $\vJh=\vJ$.
This provides a baseline without Jacobian-estimation errors. We then consider
the practical case, where the controller uses the estimate $\vJh$ obtained as
described in Section~\ref{sec:estimation}. For either case, combining
\eqref{eq:fwd} and \eqref{eq:ctrl} gives
\begin{equation}
    \dot{\ve}=-\alpha\,\vJ\vJh^{+}\ve.
    \label{eq:error_dynamics_est}
\end{equation}
Consider the Lyapunov candidate $\mathcal{L}(\ve)=\tfrac{1}{2}\norm{\ve}^{2}$.
Its derivative is $\dot{\mathcal{L}}=-\alpha\ve^{\top}\vJ\vJh^{+}\ve$, so the
error norm decreases whenever
\begin{equation}
    \ve^{\top}\vJ\vJh^{+}\ve>0.
    \label{eq:poscond}
\end{equation}

For $n\leq m$, a full-row-rank Jacobian provides local control authority in
every feature direction. In the ideal case, $\vJh=\vJ$ and
$\vJ\vJ^{+}=\mathbf{I}_n$, where $\mathbf{I}_n$ is the $n\times n$ identity
matrix, giving exponential convergence of the complete feature error. With an
estimated Jacobian, a convenient sufficient condition is
$\operatorname{Sym}(\vJ\vJh^{+})\succ0$, where
$\operatorname{Sym}(\mathbf{A})=(\mathbf{A}+\mathbf{A}^{\top})/2$ and $\succ0$ denotes positive
definiteness. In practical terms, the feature-motion directions predicted by
$\vJh$ must remain sufficiently aligned with those generated by the true
Jacobian, so that the commanded motion decreases the feature error. The case
$n=m$ is the corresponding square case.

\begin{table*}[t]
\centering
\caption{Local Stability and Convergence with an Estimated Deformation
Jacobian}
\label{tab:stability_conditions}
\footnotesize
\renewcommand{\arraystretch}{1.25}
\setlength{\tabcolsep}{7pt}

\begin{tabular}{@{}p{0.17\textwidth} p{0.34\textwidth} p{0.41\textwidth}@{}}
\toprule
\textbf{Case}
& \textbf{Sufficient local condition} & \textbf{Stability and convergence} \\
\midrule

$n\leq m$, maximum rank & $\operatorname{rank}(\vJ)=n$ and
$\operatorname{Sym}(\vJ\vJh^{+})\succ0$ & $\ve=\mathbf{0}$ is locally
exponentially stable and $\ve\to\mathbf{0}$. For $n=m$, full row rank
corresponds to the nonsingular square case. \\

$n>m$, maximum rank & $\operatorname{rank}(\vJ)=m$ and
$\operatorname{Sym}(\vJh^{+}\vJ)\succ0$ & The controllable closed-loop modes
converge exponentially, and $\ve$ approaches a residual in
$\operatorname{Null}(\vJh^{\top})$. For exact Jacobian knowledge ($\vJh=\vJ$),
this becomes the local least-squares residual in
$\operatorname{Null}(\vJ^{\top})$ and is zero when
$\ve\in\operatorname{Range}(\vJ)$. \\

Rank deficient & $\operatorname{rank}(\vJ)<\min(n,m)$, with
$\operatorname{Sym}(\vJ\vJh^{+})$ positive definite when restricted to
$\operatorname{Range}(\vJ)$ for $n\leq m$, or $\operatorname{Sym}(\vJh^{+}\vJ)$
positive definite when restricted to $\operatorname{Range}(\vJ^{\top})$ for
$n>m$ & Local exponential convergence is guaranteed only on the remaining
controllable subspace. Error components outside $\operatorname{Range}(\vJ)$
cannot be reduced locally; rank loss in $\vJh$ may remove additional corrective
directions. \\

\bottomrule
\end{tabular}

\end{table*}

For $n>m$, an arbitrary feature error cannot generally be corrected in full,
because the robot provides fewer independent motion directions than feature
coordinates. With exact Jacobian knowledge ($\vJh=\vJ$), $\vJ^{+}\ve$ is the
unique least-squares robot motion when $\vJ$ has full column rank. The matrix
$\vJ\vJ^{+}$ is then the orthogonal projector onto $\operatorname{Range}(\vJ)$:
the locally reachable component of the error is corrected, while any component
in $\operatorname{Null}(\vJ^{\top})$ remains as a least-squares residual. Here,
$\operatorname{Null}(\vJ^{\top})$ contains the feature-error directions
orthogonal to $\operatorname{Range}(\vJ)$. With an estimated Jacobian,
convergence along the available actuation directions is guaranteed by the
analogous sufficient condition $\operatorname{Sym}(\vJh^{+}\vJ)\succ0$. An error
in $\operatorname{Null}(\vJh^{\top})$ produces no commanded motion, so
estimation errors can also introduce directions in which the controller stalls.

The preceding convergence results are local, even in the favorable case of exact
Jacobian knowledge and sufficient local actuation ($m\geq n$ with full row
rank). The deformation Jacobian is a local linear model whose value and rank
generally vary with the robot and object configuration. Hence, the rank and
stability conditions above cannot be assumed to hold throughout an arbitrary
manipulation. Larger deformations may also change the object mechanics,
introduce bifurcations or contact changes, or invalidate assumptions underlying
the local model. The analysis therefore applies only within a local region where
the model and the stated conditions remain valid; when $\ve=\mathbf{0}$ is the
closed-loop equilibrium, this region defines a local basin of attraction rather
than a global convergence guarantee.

Table~\ref{tab:stability_conditions} summarizes these local conditions; setting
$\vJh=\vJ$ recovers the known-Jacobian case. The first two rows assume maximum
rank of the true Jacobian, and the last row covers rank loss.

For the discrete implementation, local convergence requires each controllable closed-loop mode to remain inside the unit circle.
For $n\leq m$, this gives $|1-\alpha\Delta t\,\lambda_i(\vJ\vJh^{+})|<1$ for all
controllable modes. For $n>m$, the corresponding condition is
$|1-\alpha\Delta t\,\lambda_i(\vJh^{+}\vJ)|<1$. Here $\lambda_i(\cdot)$ denotes the $i$th
eigenvalue of the corresponding matrix. When the Jacobian is known exactly,
$\vJh=\vJ$, and $\vJ$ has maximum rank, the relevant matrix product reduces to
the identity on the controllable directions, yielding the condition $0<\alpha\Delta t<2$.

\begin{intuitions}{Stability in Practice}
When $n\leq m$ and $\vJ$ has sufficient rank, errors can be corrected locally in
every feature direction. When $n>m$, only the component of the error within the
locally reachable feature directions, i.e., $\operatorname{Range}(\vJ)$, can be
corrected. With an estimated Jacobian, convergence further requires the
estimated mapping to preserve the corrective directions of the true Jacobian.
These guarantees remain local: rank loss, model-estimation errors, or changes in
the object mechanics can invalidate the conditions outside the local region
covered by the analysis.
\end{intuitions}

\begin{runningexample}{Closed-Loop Behavior}
Figure~\ref{fig:control} shows the shape control process with $\Delta t=1$. With
the online Broyden update \eqref{eq:broyden}, $\vJh_k$ is updated as the rod deforms and
the error converges to zero. If $\vJh_0$ is kept fixed, the estimate becomes
less representative as the deformation progresses, and the error departs from
the nominal decay. The gain comparison illustrates another consequence of the
local validity of the Jacobian estimate. For $\alpha=0.05$, the error closely
follows $(1-\alpha)^k$, whereas larger gains produce larger robot motions
between updates, so the current Jacobian estimate is applied over a larger
change in shape and becomes less accurate. At $\alpha=0.35$, the error initially increases before it converges.
\end{runningexample}

\section{Conclusion}
\label{sec}

This tutorial presented a complete data-driven Jacobian-based formulation for
shape servoing, from shape representation and error definition to deformation
Jacobian estimation, online adaptation, and closed-loop control
(Fig.~\ref{fig:loop}, Algorithm~\ref{alg:servoing}). The central idea is to
estimate directly from robot motions and measured shape changes the local
input--output relation required by the controller. This avoids requiring an
explicit mechanical model of the deformable object while retaining a compact
control formulation that can be implemented and analyzed from measured data.

The resulting guarantees are inherently local. The deformation Jacobian
describes the shape response only around the current robot--object
configuration, and its value, rank, and estimation accuracy may change as the
object deforms. Within a region where the local model and the required rank
conditions remain valid, the feedback law reduces the shape error when the
estimated inverse preserves the corrective directions of the true Jacobian
\eqref{eq:poscond}. Sufficient actuation can then provide local convergence of
the complete feature error, whereas underactuated or rank-deficient cases
generally converge only along the locally reachable feature directions and may
leave a residual error.

Part~II develops the limitations behind these assumptions in greater detail. In
particular, it examines loss of rank and poor conditioning of the deformation
Jacobian, insufficient sensitivity of the selected shape features, and changes
in the object's mechanical response. These effects are related through the
singular-value structure of the Jacobian and used to characterize loss of
control authority, loss of shape sensitivity, and mechanical instabilities. The
second part also considers extensions of the basic formulation to
three-dimensional objects, time-varying targets, or secondary control objectives.

\FundingAcknowledgment

\bibliographystyle{IEEEtran}
\bibliography{references}

\begin{thebibliography}{10}
\providecommand{\url}[1]{#1}
\csname url@samestyle\endcsname
\providecommand{\newblock}{\relax}
\providecommand{\bibinfo}[2]{#2}
\providecommand{\BIBentrySTDinterwordspacing}{\spaceskip=0pt\relax}
\providecommand{\BIBentryALTinterwordstretchfactor}{4}
\providecommand{\BIBentryALTinterwordspacing}{\spaceskip=\fontdimen2\font plus
\BIBentryALTinterwordstretchfactor\fontdimen3\font minus
  \fontdimen4\font\relax}
\providecommand{\BIBforeignlanguage}[2]{{%
\expandafter\ifx\csname l@#1\endcsname\relax
\typeout{** WARNING: IEEEtran.bst: No hyphenation pattern has been}%
\typeout{** loaded for the language `#1'. Using the pattern for}%
\typeout{** the default language instead.}%
\else
\language=\csname l@#1\endcsname
\fi
#2}}
\providecommand{\BIBdecl}{\relax}
\BIBdecl

\bibitem{hutchinson1996}
S.~Hutchinson, G.~D. Hager, and P.~I. Corke, ``A tutorial on visual servo
  control,'' \emph{IEEE Transactions on Robotics and Automation}, vol.~12,
  no.~5, pp. 651--670, 1996.

\bibitem{chaumette2006}
F.~Chaumette and S.~Hutchinson, ``Visual servo control. {Part} {I}: Basic
  approaches,'' \emph{IEEE Robotics \& Automation Magazine}, vol.~13, no.~4,
  pp. 82--90, 2006.

\bibitem{chaumette2007}
------, ``Visual servo control. {Part} {II}: Advanced approaches,'' \emph{IEEE
  Robotics \& Automation Magazine}, vol.~14, no.~1, pp. 109--118, 2007.

\bibitem{arriola2020}
V.~E. Arriola-Rios, P.~Guler, F.~Ficuciello, D.~Kragic, B.~Siciliano, and J.~L.
  Wyatt, ``Modeling of deformable objects for robotic manipulation: A tutorial
  and review,'' \emph{Frontiers in Robotics and AI}, vol.~7, 2020, {A}rt. no.
  82.

\bibitem{yin2021}
H.~Yin, A.~Varava, and D.~Kragic, ``Modeling, learning, perception, and control
  methods for deformable object manipulation,'' \emph{Science Robotics},
  vol.~6, no.~54, 2021, {A}rt. no. eabd8803.

\bibitem{cuiral2024}
I.~Cuiral-Zueco and G.~L\'opez-Nicol\'as, ``Taxonomy of deformable object shape
  control,'' \emph{IEEE Robotics and Automation Letters}, vol.~9, no.~10, pp.
  9015--9022, 2024.

\bibitem{caporali2026robotic}
A.~Caporali, I.~Cuiral-Zueco, G.~L{\'o}pez-Nicol{\'a}s, and G.~Palli, ``Robotic
  perception and manipulation of deformable linear objects: A survey,''
  \emph{The International Journal of Robotics Research}, p. 02783649261432253,
  2026.

\bibitem{shetab2022}
M.~Shetab-Bushehri, M.~Aranda, Y.~Mezouar, and E.~\"Ozg\"ur,
  ``As-rigid-as-possible shape servoing,'' \emph{IEEE Robotics and Automation
  Letters}, vol.~7, no.~2, pp. 3898--3905, 2022.

\bibitem{mcconachie2018}
D.~McConachie and D.~Berenson, ``Estimating model utility for deformable object
  manipulation using multiarmed bandit methods,'' \emph{IEEE Transactions on
  Automation Science and Engineering}, vol.~15, no.~3, pp. 967--979, 2018.

\bibitem{hosoda1994}
K.~Hosoda and M.~Asada, ``Versatile visual servoing without knowledge of true
  {J}acobian,'' in \emph{Proc. IEEE/RSJ Int. Conf. Intelligent Robots and
  Systems (IROS)}, 1994, pp. 186--193.

\bibitem{navarro2013}
D.~Navarro-Alarcon, Y.-H. Liu, J.~G. Romero, and P.~Li, ``Model-free visually
  servoed deformation control of elastic objects by robot manipulators,''
  \emph{IEEE Transactions on Robotics}, vol.~29, no.~6, pp. 1457--1468, 2013.

\bibitem{lagneau2020}
R.~Lagneau, A.~Krupa, and M.~Marchal, ``Automatic shape control of deformable
  wires based on model-free visual servoing,'' \emph{IEEE Robotics and
  Automation Letters}, vol.~5, no.~4, pp. 5252--5259, 2020.

\bibitem{berenson2013manipulation}
D.~Berenson, ``Manipulation of deformable objects without modeling and
  simulating deformation,'' in \emph{Proc. IEEE/RSJ Int. Conf. Intelligent
  Robots and Systems (IROS)}, 2013, pp. 4525--4532.

\bibitem{navarro2018}
D.~Navarro-Alarcon and Y.-H. Liu, ``Fourier-based shape servoing: A new
  feedback method to actively deform soft objects into desired {2-D} image
  contours,'' \emph{IEEE Transactions on Robotics}, vol.~34, no.~1, pp.
  272--279, 2018.

\bibitem{cherubini2020}
A.~Cherubini, V.~Ortenzi, A.~Cosgun, R.~Lee, and P.~Corke, ``Model-free
  vision-based shaping of deformable plastic materials,'' \emph{The
  International Journal of Robotics Research}, vol.~39, no.~14, pp. 1739--1759,
  2020.

\bibitem{cuiral2025time}
I.~Cuiral-Zueco and G.~López-Nicolás, ``Time consistent surface mapping for
  deformable object shape control,'' \emph{IEEE Transactions on Automation
  Science and Engineering}, vol.~22, pp. 11\,099--11\,111, 2025.

\bibitem{yang2023}
B.~Yang, C.~Sui, F.~Zhong, and Y.-H. Liu, ``Modal-graph {3D} shape servoing of
  deformable objects with raw point clouds,'' \emph{The International Journal
  of Robotics Research}, vol.~42, no.~14, pp. 1213--1244, 2023.

\bibitem{cuiral2023shapecompliant}
I.~Cuiral-Zueco, Y.~Karayiannidis, and G.~López-Nicolás, ``Contour based
  object-compliant shape control,'' \emph{IEEE Robotics and Automation
  Letters}, vol.~8, no.~8, pp. 5164--5171, 2023.

\bibitem{broyden1965}
C.~G. Broyden, ``A class of methods for solving nonlinear simultaneous
  equations,'' \emph{Mathematics of Computation}, vol.~19, no.~92, pp.
  577--593, 1965.

\end{thebibliography}

\end{document}